\documentclass[journal]{IEEEtran}

\usepackage{mmstyles}
\usepackage[linesnumbered,ruled,vlined]{algorithm2e}
\usepackage{makecell}
\usepackage{booktabs}
\usepackage{multirow}
\usepackage{graphicx}
\usepackage{pifont}
\newcommand{\cmark}{\ding{51}}%
\newcommand{\xmark}{\ding{55}}%

\SetKwInput{KwInput}{Input}
\SetKwInput{KwInitialize}{Initialize}
\SetKwInput{KwOutput}{Output}

\ifCLASSINFOpdf
\else
\fi
\begin{document}
%
\title{Improving Weakly Supervised Temporal Action Localization by Exploiting Multi-resolution Information in Temporal Domain}
%
%
%

\author{Rui~Su,
        Dong~Xu,~\IEEEmembership{Fellow,~IEEE,}
        Luping~Zhou,~\IEEEmembership{Senior Member,~IEEE,}
        and~Wanli~Ouyang,~\IEEEmembership{Senior Member,~IEEE}
\thanks{Rui Su, Dong Xu, Luping Zhou and Wanli Ouyang are with the School of Electrical and Information Engineering, The University of Sydney, NSW, Australia. Dong Xu is the corresponding author.\protect\\
E-mail: rui.su@sydney.edu.au, dong.xu@sydney.edu.au, luping.zhou@sydney.edu.au, wanli.ouyang@sydney.edu.au.}}

%
%

\markboth{Journal of \LaTeX\ Class Files,~Vol.~14, No.~8, August~2015}%
{Shell \MakeLowercase{\textit{et al.}}: Bare Demo of IEEEtran.cls for IEEE Journals}
%



\maketitle

\begin{abstract}
Weakly supervised temporal action localization is a challenging task as only the video-level annotation is available during the training process. 
To address this problem, we propose a two-stage approach to fully exploit multi-resolution information in the temporal domain and generate high quality frame-level pseudo labels based on both appearance and motion streams.
Specifically, in the first stage, we generate reliable initial frame-level pseudo labels, and in the second stage, we iteratively refine the pseudo labels and use a set of selected frames with highly confident pseudo labels to train neural networks and better predict action class scores at each frame.
We fully exploit temporal information at multiple scales to improve temporal action localization performance.
Specifically, in order to obtain reliable initial frame-level pseudo labels, in the first stage, we propose an Initial Label Generation (ILG) module, which leverages temporal multi-resolution consistency to generate high quality class activation sequences (CASs), which consist of a number of sequences with each sequence measuring how likely each video frame belongs to one specific action class. 
%
In the second stage, we propose a Progressive Temporal Label Refinement (PTLR) framework.
In our PTLR framework, two networks called Network-OTS and Network-RTS, which are respectively used to generate CASs for the original temporal scale and the reduced temporal scales, are used as two streams (\ie, the OTS stream and the RTS stream) to refine the pseudo labels in turn.
By this way, the multi-resolution information in the temporal domain is exchanged at the pseudo label level, and our work can help improve each stream (\ie, the OTS/RTS stream) by exploiting the refined pseudo labels from another stream (\ie, the RTS/OTS stream).
Comprehensive experiments on two benchmark datasets THUMOS14 and ActivityNet v1.3 demonstrate the effectiveness of our newly proposed method for weakly supervised temporal action localization.
\end{abstract}


%
\IEEEpeerreviewmaketitle

\section{Introduction}
%
%
%
%
\IEEEPARstart{R}{ecently}, the increasing number of untrimmed videos raises the demand for analyzing human behaviours in these videos.
Accordingly, the existing works~\cite{zeng2019graph,chao2018rethinking,gao2018ctap,lin2018bsn,lin2019bmn,wang2016temporal} have made impressive progress for temporal action localization, which aims to locate the temporal starting and ending frames for detecting the actions of interest in the untrimmed videos.
However, it is often required to annotate the starting and ending positions in order to train a robust deep model for the temporal action localization task, which could be very expensive.
Moreover, temporal boundary annotation often tends to be subjective, thus could be misleading.
Therefore, weakly supervised temporal action localization, which significantly reduces human labelling costs, has gained increasing research interest from the research community.

For weakly supervised temporal action localization, its goal is also to localize the temporal boundaries of actions in untrimmed videos.
However, instead of using strong supervision (\ie, temporal boundary annotation indicating the starting and ending frames of an action instance), only video-level action labels are available as weak supervision.
This task can be formulated as how to predict the action label for each individual frame.
To predict the frame-level action class labels, most existing works~\cite{lee2020background, nguyen2019weakly,nguyen2018weakly} generate the so-called class activation sequences (CASs) for the whole video by using multiple instance learning.
On the other hand, the frame-level pseudo action labels can also be readily obtained from the generated CASs.
Based on these pseudo labels, we can also train a classification network to predict the frame-level action class labels.
However, without using ground-truth frame-level annotation as strong supervision (\ie, temporal boundary annotation), a deep network trained based on weak supervision often generates CASs that only cover the most discriminative frames or incorrectly detected background frames, which cannot meet the requirement to generate reliable frame-level pseudo action class labels for training a robust frame-level classification network.

To address this problem, we propose to improve the quality of the generated CASs by exploiting more constraints, which can be provided by the temporal multi-resolution consistency.
%
%
%
%
%
It is obvious that the contents of a video are almost the same regardless of its temporal resolution.
Intuitively, the generated CASs should not be significantly affected by the temporal resolution of the input video.
%
However, we observe that the CASs generated by using the same input video from different temporal resolutions are not always consistent.
Inspired by this observation, we propose a temporal multi-resolution consistency loss to constrain the CASs generated by using the same input video at different temporal resolutions to be consistent to each other, so that the quality of the generated CASs is enhanced. 
The improved CASs will be then used to obtain better frame-level pseudo action labels.
This is our first strategy to exploit the temporal multi-resolution information.
We also observe that the input videos at different temporal resolutions can capture fast or slow motion information, and integration of all types of motion information could benefit the action class score prediction.
Specifically, the pseudo labels produced based on the CASs from the original resolution videos and the temporally scaled videos also contain complementary information.
%
%
Moreover, we also observe that the frames within a short period of time often have the same action class labels, so that we can upsample the shorter CASs produced from the shorter duration input videos to the CASs produced from the original video. 
Inspired by this observation, we propose a two-stream approach to make use of complementary information from the pseudo labels produced from the CASs that are respectively generated based on the original duration videos and the reduced duration videos, and exchange the pseudo labels from two streams to progressively expand the training set by using more frames with more reliable pseudo labels.
This is our second strategy to exploit the multi-resolution information in the temporal domain.

Based on our first strategy, in this work, we first propose an Initial Label Generation (ILG) module to exploit the temporal multi-resolution consistency by identifying the action frames with more accurate pseudo labels.
Specifically, our ILG module builds upon a two-stream framework~\cite{simonyan2014two} and the recent weakly supervised temporal action localization framework~\cite{lee2020background}, in which we generate the CASs for each stream (\ie, the RGB/flow stream).
For each stream, we use two branches with the network in each branch respectively copes with the original videos and the temporally scaled videos and also propose a temporal multi-resolution consistency loss to enforce the CASs generated from two branches to be consistent.
Additionally, a cross-stream consistency loss is also imposed to constrain the CASs generated from the two streams (\ie, the RGB and flow streams) to be consistent, by which complementary information between the RGB and flow streams could also be utilized.
%
%
By enforcing the temporal multi-resolution consistency and the cross-stream consistency, we make use of extra constraints to enable our ILG module to generate better CASs with higher action class prediction probabilities.
The CASs generated from the two streams are combined to obtain the pseudo action labels, which are then used to train an initial classification network in the second stage.

To further improve the weakly supervised temporal action localization performance, based on our second strategy, we also introduce the Progressive Temporal Label Refinement (PTLR) framework, which consists of two networks, in which the Network-OTS is to predict the frame-level action class scores based on original videos, and the Network-RTS is to predict the frame-level action class scores based on the features from the temporally scaled videos.
Our PTLR framework exploits multi-resolution information in the temporal domain at the pseudo label level between these two networks in order to iteratively refine the pseudo labels.
%
%
The Network-OTS is first trained by using the initial pseudo labels generated from the ILG module and outputs the CASs, which can be used to generate more reliable pseudo labels as supervision to train the Network-RTS.
%
%
%
In Network-RTS, a set of temporal convolution operations are used to extract the features from the temporally scaled videos and generate CASs at multiple temporal resolutions, which are all upsampled and then combined to output the CAS at the original temporal resolution.
In this way, more reliable pseudo labels can be obtained to retrain the Network-OTS.
%
This refinement process is repeated by using the Network-OTS and the Network-RTS in turn to refine the pseudo labels alternatively.
In this way, better weakly supervised temporal action localization performance can be achieved.

Our contributions can be summarized as follows:
\begin{itemize}
    \item We propose a weakly supervised temporal action localization framework that is able to generate high-quality frame-level pseudo action labels by exploiting temporal multi-resolution information in its Initial Label Generation (ILG) module and Progressive Temporal Label Refinement (PTLR) module.
    \item In the ILG module, we propose a temporal multi-resolution consistency loss and a cross-stream consistency loss to constrain the generated CASs so that more reliable initial pseudo labels can be obtained.
    \item In the PTLR module consisting of two networks called Network-OTS and Network-RTS, we exploit the temporal multi-resolution information by iteratively refining the pseudo labels to train better networks, and use the enhanced networks to further improve the quality of pseudo labels.
    \item Comprehensive experiments on two benchmark datasets THUMOS14~\cite{jiang2014thumos} and ActivityNet v1.3~\cite{caba2015activitynet} demonstrate our approach outperforms the state-of-the-art methods for the weakly supervised temporal action localization task.
\end{itemize}


\section{Related Work}

\subsection{Video Action Classification}
Video representation learning is widely studied in video action analysis, especially in the field of action recognition. 
Recent works~\cite{xie2018rethinking,feichtenhofer2016convolutional,simonyan2014two,wang2016temporal,carreira2017quo,tran2015learning,qiu2017learning} have achieved impressive performance improvement on action recognition.
These works can be roughly categorized into two types: (1) The works in~\cite{feichtenhofer2016convolutional, simonyan2014two, wang2016temporal} consider both appearance and motion information as the discriminative clues for action recognition.
Building upon a two-stream framework, their methods separately extract features from RGB images and optical flow maps to represent appearance and motion information, respectively.
The RGB and flow features are then combined to form a new feature to represent the video in order to take advantage of complementary information between these two types of clues.
Unlike in~\cite{simonyan2014two} and~\cite{wang2016temporal}, instead of fusing the features from two streams in a late fusion fashion, the work in~\cite{feichtenhofer2016convolutional} also investigates how to fuse the two-stream features in the early fusion way.
(2) The works in ~\cite{xie2018rethinking, carreira2017quo, tran2015learning,qiu2017learning} apply the 3D convolutional operations to extract spatio-temporal representations.
For example, in~\cite{carreira2017quo}, they extend the 2D convolutional operations in~\cite{szegedy2015going} to the 3D convolutional operations and simultaneously model spatio-temporal information from the videos.
In addition to address the video classification problems, all the aforementioned approaches can also be used to extract spatio-temporal representation.
In our work, we use the I3D features~\cite{carreira2017quo} for the weakly supervised temporal action localization task.


\subsection{Fully Supervised Temporal Action Localization}

In the fully supervised temporal action localization task, strong annotation at the temporal domain (\ie, the starting and ending frames are provided) is used to train the temporal action localization models, which are then used to detect the starting and the ending points of action instances in untrimmed videos. 
The early works~\cite{shou2016temporal, yuan2016temporal} use slide windows with different temporal durations to generate the proposals and then use the classification network to predict action class labels for each proposal.
There are two lines of works for fully supervised temporal action localization: (1) In~\cite{ma2016learning, yuan2017temporal, zhao2017temporal, shou2017cdc}, they extract frame-level features and use the frame-level features to generate frame-level action class labels, which are then used to decide the temporal boundaries of action instances.
For example, the work in~\cite{shou2017cdc} uses the 3D-CNN to downsample the clip features and then upsample the features to their original scales in the temporal domain in order to obtain frame-level action scores to determine the starting and the ending points of action instances.
(2) The works in~\cite{chao2018rethinking,zeng2019graph} follow the two-stage framework for object detection, in which a region proposal network is first applied to generate a set of proposals and then the regression and classification networks are used to refine the temporal boundaries and predict the action class labels.
All aforementioned works for fully supervised temporal action localization use strong annotation at the temporal domain during the training process, while we aim to use only weak supervision (\ie, the video-level labels) for temporal action localization.


\subsection{Weakly Supervised Temporal Action Localization}

Similar as the fully supervised temporal action localization task, weakly supervised temporal action localization also aims to detect the temporal boundaries for action of interest in untrimmed videos.
However, instead of using strong supervision, only weak supervision (\ie, the video-level labels) are available during the training process.
The existing works~\cite{nguyen2018weakly, nguyen2019weakly,lee2020background,xu2019segregated} treat it as a multiple instance learning (MIL) problem., in which the labels for a bag of instances rather than each individual instance are employed.
The works in~\cite{nguyen2019weakly} and~\cite{lee2020background} generate the so-called class activation sequences (CASs) to obtain the proposals, and both works consider how to suppress the activation for background frames in order to generate CAS with high quality.
Additionally, the method in~\cite{shou2018autoloc} proposes an Outer-Inner-Contrastive Loss to predict the temporal boundaries of action instances and the work in~\cite{liu2019completeness} predicts the completeness of actions by employing a multi-branch architecture.
However, the aforementioned works do not consider the cross-stream consistency and temporal multi-resolution consistency when generating CAS.
In contrast, our work employs the cross-stream consistency loss and the temporal multi-resolution consistency loss as self-supervision information to generate better CASs.
%


\subsection{Temporal Multi-resolution Information}
Recently, the work in~\cite{feichtenhofer2019slowfast} exploits the temporal multi-resolution information for video analysis. 
It captures the spatial information by using input videos at low temporal resolution, and captures motion information by using input videos at high temporal resolution. 
The temporal multi-resolution information is then combined to improve the performance for both action classification and spatial action detection tasks.
In contrast to~\cite{feichtenhofer2019slowfast}, our work exploits the temporal multi-resolution information for weakly supervised temporal action localization instead of the action classification and spatial action detection tasks as in~\cite{feichtenhofer2019slowfast}. In addition, we exploit the temporal multi-resolution information by using a new loss function and exchanging the pseudo labels, which is also different from~\cite{feichtenhofer2019slowfast}.


\begin{figure*}[!t]
\includegraphics[width=\linewidth]{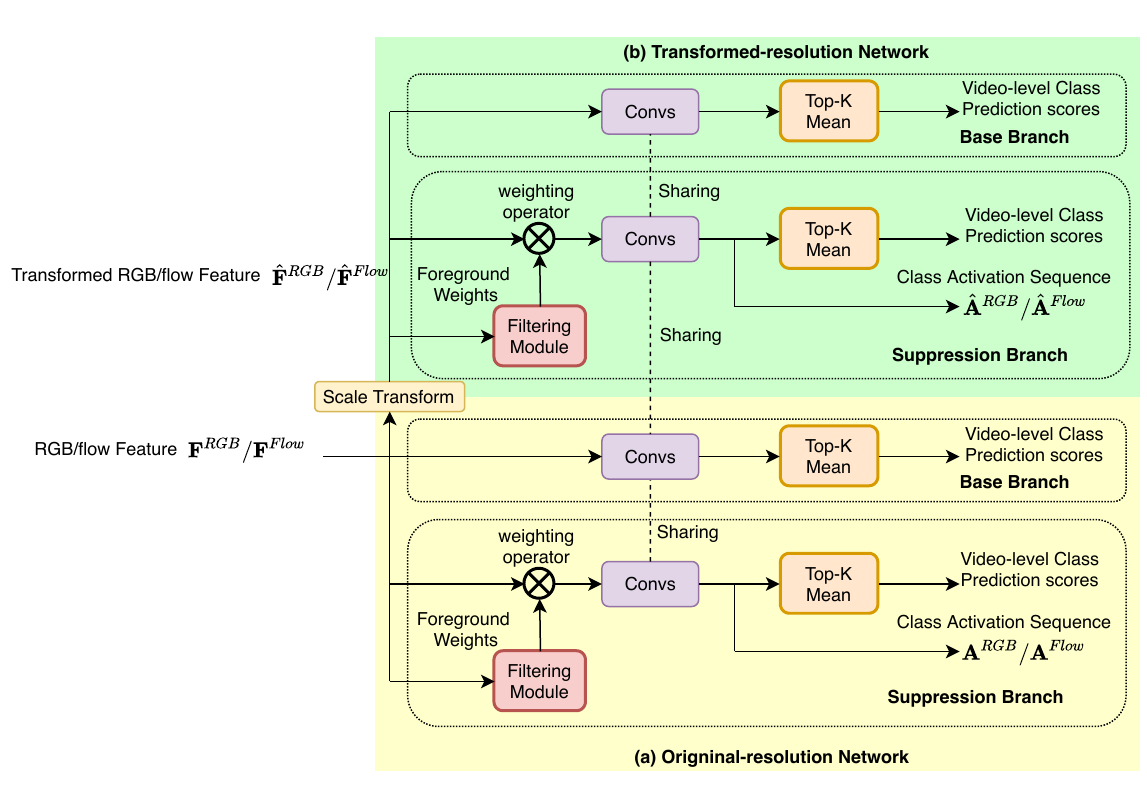}
\caption{The overall framework of our Initial Label Generation (ILG) module. Our framework consists of two networks: (a) Original-resolution Network, (b) Transformed-resolution Network. The temporal scale transform operation is applied on the RGB/flow feature $\mF^{\text{RGB}}/\mF^{\text{Flow}}$ to obtain the transformed RGB/flow feature $\hat{\mF}{}^{\text{RGB}}/\hat{\mF}{}^{\text{Flow}}$. The networks in (a) and (b) take $\mF^{\text{RGB}}/\mF^{\text{Flow}}$ and $\hat{\mF}{}^{\text{RGB}}/\hat{\mF}{}^{\text{Flow}}$ as the input and use Bas-Net~\cite{lee2020background} to generate CASs $\mA^{\text{RGB}}/\mA^{\text{Flow}}$ and $\hat{\mA}{}^{\text{RGB}}/\hat{\mA}{}^{\text{Flow}}$, respectively. and note each network also consists of a base branch and a suppression branch.}
\label{fig:ILG}
\end{figure*}

\section{Our Approach}

In this section, we first briefly review the Bas-Net~\cite{lee2020background} in Section~\ref{sec:bas-net} and introduce the overall framework of our method in Section~\ref{sec:overview}, and then elaborate our Initial Label Generation (ILG) and Progressive Temporal Label Refinement (PTLR) modules in Section~\ref{sec:ILG} and Section~\ref{sec:PTLR}, respectively.

\subsection{Bas-Net} \label{sec:bas-net}

In this section, we briefly review the Bas-Net~\cite{lee2020background} for the weakly supervised temporal action localization task, which is used as the backbone in our method to generate CASs.

\textbf{Feature Extraction}~~
%
Given an untrimmed video, the visual features that represent the whole video are extracted from the pretrained feature extractors for both the RGB and the flow streams. 
Take the RGB stream as an example. 
For a RGB sequence $V^{RGB}$ with $T$ frames, a pretrained I3D~\cite{carreira2017quo} feature extractor is applied to extract the feature vector $\vf^{RGB}_{t} \in R^{D}$ , where $D$ is the feature dimension for each frame, and the frame index $t=1,2,...,T$. 
The feature vectors $\{\vf^{RGB}_{t}\}_{t=1,2,...,T}$ are then stacked along the temporal dimension to form the RGB feature matrix $\mF^{RGB} = [\vf^{RGB}_1,\vf^{RGB}_2,...,\vf^{RGB}_T] \in R^{D \times T}$ for the whole video.
Similarly, the flow feature matrix $\mF^{Flow}$ for the whole video can be extracted from the flow sequence $V^{Flow}$.
Afterwards, the RGB feature $\mF^{RGB}$ and the flow features $\mF^{Flow}$ are fused to produce the fused two-stream feature $\mF^{\text{fused}}$ in an early fusion fashion.

\textbf{Base Branch}~~
The base branch applies 1D convolutional operations on the fused feature $\mF^{\text{fused}}$ and produces the frame-level action class scores, which form a class action sequence (CAS).
The generated CAS is then aggregated to obtain video-level class scores $\mP^{\text{base}}$ by using the top-k mean operation, which is supervised by the ground-truth video-level action class labels.
Considering that each video contains the background frames, all training videos are labelled as positive for the background class (\ie, the video-level label for the video in the base branch is denoted as $\vy^{\text{base}} = [y_1,y_2,...,y_C,1] \in R^{C+1}$, where $C$ is the total number of action classes, and $y_c=1|_{c=1,2,...,C}$ if the video contains the $c^{\text{th}}$ action class).
As all the videos contain the background class, the generated CAS will also be biased towards the background class.
A suppression branch is thus employed to address this issue.

\textbf{Suppression Branch}~~
In the suppression branch, a filtering module is applied to take the fused feature $\mF^{\text{fused}}$ as the input and output the foreground weights, which are then used to weigh the input feature $\mF^{\text{fused}}$ and suppress the effect from the background frames before being fed into the convolutional layers.
It is noted that the convolutional layers used in both the base branch and the suppression branch share the parameters.
The video-level class scores for the suppression branch can be obtained by the same process as described in the base branch.
However, to suppress the effect from the background frames, all training videos are labelled as negative for the background class (\ie, the video-level label for the suppression branch is denoted as $\vy^{\text{supp}}=[y_1,y_2,...,y_C,0]\in R^{C+1}$).
The generated CASs from the suppression branch are then used to produce the action instances by discarding the frames with action prediction scores lower than a pre-defined threshold and grouping the consecutive frames in the remaining frames.

\textbf{Objective function}~~
To train the Bas-Net, the binary cross-entropy loss for each class is used in the objective functions for the base branch and the suppression branch, which are denoted as $L_{\text{base}}$ and $L_{\text{supp}}$, respectively.
%
%
The whole Bas-Net is trained jointly by minimizing the overall loss function as follow:
\begin{equation} \label{eq:1}
    L_{\text{bas-net}} = L_{\text{base}} + L_{\text{supp}} + L_{\text{norm}},
\end{equation}
where $L_{\text{norm}}$  is the L1 normalization term of the foreground weights obtained from the filtering module in the suppression branch.

\subsection{Overview of Our Approach} \label{sec:overview}
%
Different from the Bas-Net, which outputs the frames containing the action instances directly from the CAS, we first use the CAS to generate the initial frame-level pseudo labels, and then use the generated initial pseudo action class labels to train a temporal action localization network.
Specifically, our work consists of an Initial Label Generation (ILG) module and a Progressive Temporal Label Refinement (PTLR) module. The ILG module takes an untrimmed video as the input and outputs the initial frame-level pseudo action class labels by generating better CASs, in which the temporal multi-resolution consistency and the cross-stream consistency constraints are additionally enforced. 
The initial pseudo labels are then progressively refined in our Progressive Temporal Label refinement (PTLR) module , which are also used as better supervision to train an improved temporal action localization network.


\subsection{Initial Label Generation (ILG)} \label{sec:ILG}

%
%
It is well known that both appearance and motion clues are important for the weakly-supervised temporal action localization task. 
In order to take advantage of both clues, our ILG module builds upon a two-stream Bas-Net framework, where both the RGB (appearance) and the flow (motion) streams use the same network structure as in Bas-Net, which consists of a base branch and a suppression branch, but the networks from the two streams do not share the parameters.
Figure~\ref{fig:ILG} shows our Initial Label Generation (ILG) module, in which the RGB stream and the flow stream have the same network structure but they take the RGB feature and the flow feature as the input, respectively.
The details of our ILG module are introduced below.

\textbf{CAS Generation}~~
In our ILG module, Bas-Net~\cite{lee2020background} is applied to generate CASs.
We first extract the RGB/flow features $\mF^{RGB}$ and $\mF^{Flow}$ as described in Section~\ref{sec:bas-net}.
However, instead of combining the two-stream features in an early fusion fashion as in~\cite{lee2020background}, the RGB and flow features $\mF^{RGB}$ and $\mF^{Flow}$ are put into two separate Bas-Net networks to generate the CASs $\mA^{RGB}$ and $\mA^{Flow} \in R^{(C+1) \times T}$ for the RGB stream and the flow stream, respectively, where $C$ is the number of action classes.
%
%
%
We use the CASs from the suppression branch of the Bas-Net as the output as its action class prediction scores are better than those from the base branch.
%

\begin{figure*}[!t]
 \centering
\includegraphics[width=0.9\linewidth]{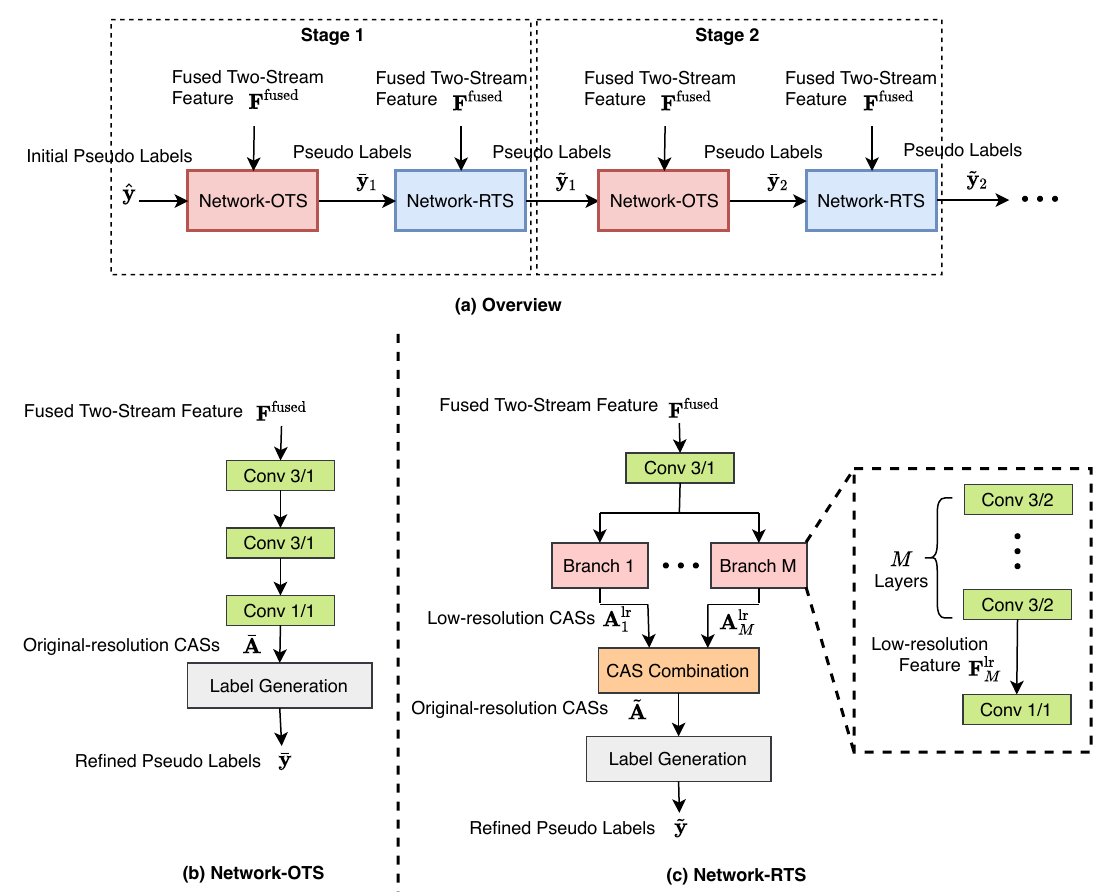}
\caption{(a)The overall structure of our Progressive Temporal Label Refinement (PTLR) framework. 
Our PTLR framework consists of multiple stages. In each stage, the Network-OTS/Network-RTS is trained by using the selected frames with the corresponding pseudo labels produced from the Network-RTS/Network-OTS in the previous iteration and then output the refined the pseudo labels, which are then used as supervision for training the Network RTS/Network-OTS in the next iteration. 
(b) (c) The details of our Network-OTS and Network-RTS in our PTLR framework. 
In the Network-OTS, it takes the fused two-stream feature $\mF^{\text{fused}}$ as the input and generates the frame-level CAS $\bar{\mA}$ in order to produce the pseudo labels $\bar{\vy}$. The selected frames together with their pseudo labels $\bar{\vy}$ are then used to train the Network-RTS. The Network-RTS consists of $M$ branches. 
For the $m^{\text{th}}$ branch, it stacks $m$ convolutional layers and take the fused feature $\mF^{\text{fused}}$ as the input to extract the low-resolution feature $\mF^{\text{lr}}_m$, which is used to generate the low-resolution CAS $\mA^{\text{lr}}_m$. Taking the low-resolution CASs generated from all branches as the input, the CAS combination block temporally upsamples these CASs and then averaged them to produce the original-resolution CAS $\tilde{\mA}$ so that the refined new pseudo labels $\tilde{\vy}$ can be generated. The selected frames with their correspoinding pseudo labels $\tilde{\vy}$ are then used to retrain the Network-OTS. Note that $n/r$ in the conv block indicates the kernel size $n$ and the stride $r$ used in the conv block.} 
\label{fig:PTLR}
\end{figure*}

\textbf{Label Generation}~~
After CAS generation, we use the weighted-average operation to combine $\mA^{RGB}$ and $\mA^{Flow}$ and produce the final CAS $\mA \in R^{(C+1) \times T}$.
%
%
%
In this work, the weights for combining $\mA^{RGB}$ and $\mA^{Flow}$ are set to 1 and 1.5, respectively, in which we assign higher weights to the motion clue.
%
After $\mA$ is produced, a threshold is applied so that the frames with high action class scores are labelled as the foreground samples related to one specific action class and the frames with low action class scores are labelled as the background samples.
Specifically, let the binary vector $\hat{{\vy}}{}_{t} \in R^{C+1}$ denote the pseudo label for the $t^{\text{th}}$ frame with its element $\hat{y}_{t,c}=\{0,1\}$ indicating whether the $t^{\text{th}}$ frame belongs to the $c^{\text{th}}$ action class. 
For the $t^{\text{th}}$ frame, if its $c^{\text{th}}$ class score $\mA_{t,c}$ is larger than the pre-defined high threshold (\ie, we use 0.7 in this work), the $t^{\text{th}}$ frame is labelled as the action class $c$ with $\hat{y}{}_{t,c}=1$ and $\hat{y}{}_{t,i}=0|_{\forall i\neq c}$, and the frame index $t$ is added into an index set of selected frames $\cT$.  
Otherwise, if $\sum_{c=1}^{C} \mA_{t,c}$ is less than the pre-defined low threshold (\ie, we use 0.1 in this work), the $t^{\text{th}}$ frame is labelled as the background class with $\hat{y}{}_{t,0}=1$ and $\hat{y}{}_{t,i}=0|_{\forall i\neq 0}$, and the frame index $t$ is also added into an index set of selected frames $\cT$.
Only the frames in $\cT$ together with the corresponding pseudo labels will be used as the training samples in our PTLR framework in the second stage, and other unselected samples will be discarded.

As the initial pseudo labels are generated from the CAS, the quality of the CAS is critical.
In this work, we impose two additional constraints (\ie, the cross-stream consistency constraint and the temporal multi-resolution consistency constraint) to further improve the quality of CAS.
The details are described as follows.

\textbf{Cross-stream Consistency Loss}~~
%
Given an untrimmed video, since the appearance and the motion clues correspond to the same video content, the predicted frame-level action class scores by using either clue should also be similar to each other.
To this end, we propose a new constraint so that the CAS generated from the RGB stream should be similar with that from the flow stream, which is formulated as a cross-stream consistency loss measuring the L1 distance between $\mA^{RGB}$ and $\mA^{Flow}$ as follows:
\begin{equation}
    \label{eq:2}
    L_{csc} = \left\|\mA^{RGB}-\mA^{Flow}\right\|_1.
\end{equation}
%
This cross-stream consistency loss is added into the objective of the ILG module (\ie, Eq.~(\ref{eq:5})) for joint optimization.
It also exploits the complementary information between the RGB and the flow streams, which allows the information from one stream (\ie, RGB/flow stream) to help train a better model for another stream (\ie, flow/RGB stream). 
%

\textbf{Temporal multi-resolution Consistency Loss}~~
In our ILG module, we additionally exploit temporal multi-resolution information by further enforcing the consistency between the CASs from different temporal scales. 
Given a feature $\mF$ of an untrimmed video and a temporally scaled feature $\hat{\mF}=ST(\mF, s)$, where linear interpolation is used as the temporal scale transform operation $ST(\cdot,s)$ on $\mF$ and $s$ is the transform scale factor, intuitively, the CAS $\hat{\mA}$ generated by using the scale transformed feature $\hat{\mF}$ should be consistent with the CAS after temporal scale transform (\ie, $ST(\mA,s)$), where the CAS $\mA$ is generated by using the original feature $\mF$. 
As a result, in addition to the cross-stream consistency loss, we also propose a temporal multi-resolution consistency loss to enforce the consistency between CASs from different temporal resolutions. 

Specifically, in each stream, we apply the scale transform on the input feature $\mF$ and add an extra branch to generate the CAS by using the transformed feature $\hat{\mF}$, and then constrain the L1 distance between the generated CAS from the transformed-resolution network and the scale transformed CAS from the original-resolution network.
Take the RGB stream as an example for illustration.
The initial RGB feature $\mF^{RGB}$ is rescaled to form the new feature $\hat{\mF}{}^{RGB}=ST(\mF,s)$ by applying the scale transform $ST(\cdot,s)$ with a random scale factor $s$ ranging from 0.5 to 2.
The transformed RGB feature $\hat{\mF}{}^{RGB}$ is then fed into the Bas-Net for CAS generation, and then we generate the CAS $\hat{\mA}{}^{RGB}$ from the transformed-resolution network.
The temporal multi-resolution consistency loss for the RGB stream can be formulated as:
\begin{equation}
    \label{eq:3}
    L^{RGB}_{sc} = \left\|ST(\mA^{RGB},s) - \hat{\mA}{}^{RGB}\right\|_1.
\end{equation}
Similarly, the transformed flow feature $\hat{\mF}{}^{Flow}$ and the corresponding CAS $\hat{\mA}{}^{Flow}$ from the transformed-resolution network can be generated.
Then we have the following loss function:
\begin{equation}
    \label{eq:4}
    L^{Flow}_{sc} = \left\|ST(\mA^{Flow},s) - \hat{\mA}{}^{Flow}\right\|_1.
\end{equation}
The temporal multi-resolution consistency loss can provide additional constraints between the original and temporally transformed CASs, which also improves the generalization capability of our ILG module to handle the input videos with various temporal scales.

\textbf{Joint Training}~~
To train our ILG module, we first apply the overall loss function of Bas-Net~\cite{lee2020background} (\ie, Eq.~(\ref{eq:1})) for each stream to obtain the two loss terms $L_{\text{bas-net}}^{RGB}$ and $L_{\text{bas-net}}^{Flow}$.
We then add these two loss terms with our newly proposed cross-stream consistency loss and temporal multi-resolution consistency loss to jointly train our ILG network:
\begin{equation}
    \label{eq:5}
    L_{\text{total}}=L_{\text{bas-net}}^{RGB} +L_{\text{bas-net}}^{Flow} + \alpha L_{csc} + \beta L_{sc}^{RGB} + \gamma L_{sc}^{Flow},
\end{equation}
where $\alpha$, $\beta$, and $\gamma$ are the hyperparameters to balance different terms.
In this work, we empirically set $\alpha=\beta=\gamma=0.5$.



\subsection{Progressive Temporal Label Refinement} \label{sec:PTLR}

After the initial pseudo labels are produced by our ILG module, we can use them as supervision to train a neural network to predict the frame-level action class confident scores and output the CASs, and then use different thresholds to decide the starting and the ending points of action instances.
%
However, as mentioned above, by using the thresholding strategy, only discriminative foreground samples (\ie, the frames with high action class prediction scores) are used in the training process and less discriminative foreground samples are discarded.
As a result, it is hard to accurately predict the frame-level action class labels by directly using the samples with the initial pseudo labels.

To this end, we propose a new Progressive Temporal Label Refinement (PTLR) framework, which takes advantage of the complementary temporal multi-resolution information by iteratively refining the pseudo labels so that they can be used to train a neural network for more accurate frame-level action class prediction. 
As shown in Figure~\ref{fig:PTLR}, our PTLR framework takes $\mF^{\text{fused}}\in R^{2D\times T}$ that concatenates $\mF^{\text{RGB}}$ and $\mF^{\text{flow}}$ as the input, and consists of two networks, in which the Network-OTS generates CASs from the features of original duration videos and the Network-RTS generates CASs from the features of multiple reduced duration videos.
The generated CASs from both networks are used to produce more reliable pseudo labels, and we can use the newly produced pseudo labels from one network (\ie, the Network-OTS/Network-RTS) to retrain another network (\ie, the Network-RTS/Network-OTS) and progressively improve the temporal action localization performance.
Our implementation details of the Network-OTS and the Network-RTS are introduced as follows.

\textbf{Network-OTS}~~
The architecture of our Network-OTS is shown in Figure~\ref{fig:PTLR}(b).
We first pass the input feature $\mF^{\text{fused}}$ to two 1D convolutional layers with the kernel size of 3 to increase the temporal reception field.
On top of this operation, we then apply a 1D convolutional operation with the kernel size of 1 to generate the frame-level action class scores $\bar{\mA} \in R^{(C+1)\times T}$. 
In order to train the Network-OTS, we optimize the loss function calculated over all selected training frames with their indexes in $\cT$, which is defined as follows:
\begin{equation}
    L_{\text{ots}} = \frac{1}{N_{\text{in}}}\sum_{t\in\cT}\sum_{c=0}^{C}\tilde{y}{}_{t,c}log(\bar{\mA}_{t,c}),
\end{equation}
where $N_{\text{in}}$ is the total number of samples in $\cT$, and $\tilde{y}{}_{t,c}$ is the pseudo labels for action class $c$ at $t^{\text{th}}$ frame produced from the Network-RTS.
Note that when the Network-OTS is first trained, instead of using the pseudo labels produced from the Network-RTS, we assign the initial pseudo label produced from our ILG framework $\hat{y}{}_{t,c}$ as the pseudo label for action class $c$ at $t^{\text{th}}$ frame.
In the training stage, $\bar{\mA}$ is optimized before generating new pseudo labels.
Specifically, we denote $\bar{\vy}_{t} \in R^{C+1}$ as the newly generated pseudo label for the $t^{\text{th}}$ frame of an untrimmed video. 
When $\bar{\mA}_{t,c}$ is larger than a pre-defined threshold $\theta_p$, we label the $t^{\text{th}}$ frame as a foreground sample with the action class label c and assign $\bar{y}_{t,c}=1$ and $\bar{y}_{t,i}=0|_{\forall i\neq c}$, and the frame index $t$ is added into an index set $\cT^{\text{ots}}$.
Similarly, for the $t^{\text{th}}$ frame, if $\bar{\mA}_{t,0}$, (\ie, the background class score ) is larger than a threshold $\theta_p$, we label the $t^{\text{th}}$ frame as a background sample and set $\bar{y}_{t,0}=1$ and $\bar{y}_{t,i}=0|_{\forall i\neq 0}$, and the frame index $t$ is also added into the index set $\cT^{\text{ots}}$.
We then update the initial pseudo label $\hat{\vy}_t$ and $\cT$ with $\bar{\vy}_t$ and $\cT^{\text{ots}}$, respectively.

\textbf{Network-RTS}~~
%
In our Network-RST, we reduce the temporal resolution of the input videos with a scale factor and generate the low-resolution CASs based on the reduced resolution videos.
%
In this work, we apply the multi-branch structure in our Network-RTS to generate CASs with  different reduced temporal resolutions in order to better exploit temporal multi-resolution information.
As shown in Figure~\ref{fig:PTLR}(c), our Network-RTS consists of $M$ branches, with each branch corresponding to a different reduced temporal resolution.
For example, for the $m^{\text{th}}$ branch, we reduce the temporal resolution of the input video with a scale factor of $2^m$ and generate the corresponding low-resolution CASs $\mA^{\text{lr}}_m \in R^{(C+1)\times (T/k_j)}$, where $m=1,2,...,M$.
Specifically, we first take the fused feature $\mF^{\text{fused}}$ as the input and apply a 1D convolutional layer with the kernel size of 3 on top of the input.
Then the output feature is fed to multiple branches to generate the features at different temporal resolutions with different scaling factors.
For the $m^{\text{th}}$ branch, we apply $m$ 1D convolutional layers with the stride of 2 and the kernel size of 3 to reduce the temporal dimension of the feature with the scale factor $2^m$.
%
Finally, we can apply a 1D convolutional operation with the kernel size of 1 on top of the output feature to generate the low-resolution CASs $\mA^{\text{rl}}_m$.
In order to train our PTLR framework, for the $m^{\text{th}}$ branch, if the temporal index $t$ of the low-resolution CASs $\mA^{\text{rl}}_m$ satisfies the condition $t\times 2^m \in \cT$, we add $t$ into the training index set $\cT^{m}$, and its corresponding pseudo label is defined as $\bar{\vy}{}^{m}_t=\bar{\vy}{}_{t\times 2^m}$.
We then define the loss function as follows:
\begin{equation}
    L_{\text{rst}}=\frac{1}{M}\sum_{m=1}^{M}\frac{1}{N_{\text{in}}^{m}}\sum_{t \in \cT^m}\sum_{c=0}^{C}\bar{y}_{t,c}^m log(\tilde{\mA}_{t,c}),
\end{equation}
where $N_{in}^m$ is the total number of indexes in $\cT^m$ for the $m^{\text{}th}$ branch.

\begin{algorithm}[t]
\label{algo:PTLR}
\small
\SetAlgoLined
\caption{Our proposed PTLR framework}
\KwInput{The fused feature set $\cF^{\text{fused}}$ extracted from the input videos, the selected training index set $\cT$ the corresponding initial pseudo label set $\hat{\cY}$, and the total number of stages $N$}
\KwInitialize{Stage $n=0$}
\While{$n<N$}{
Train the Network-OTS with the features $\cF^{\text{fused}}$ extracted from the input videos, the selected training index set $\cT$ and their corresponding pseudo labels in the set $\hat{\cY}$\\
Generate the original-resolution CASs with the trained Network-OTS\\
Produce the new pseudo label set $\bar{\cY}$ and the new index set $\cT^{\text{ots}}$ from the predicted original-resolution CASs\\
Update $\hat{\cY} \leftarrow \bar{\cY}$, ${\cT} \leftarrow {\cT}{}^{\text{ots}}$\\
Train the Network-RTS with the features $\cF^{\text{fused}}$ extracted from the input videos, the selected training index set $\cT$ and their corresponding pseudo labels in the set $\hat{\cY}$\\
Generate the low-resolution CASs with the trained Network-RTS\\
Temporally upsample the low-resolution CASs to the original resolution\\
Produce the new pseudo label set $\tilde{\cY}$ and the new index set $\cT^{\text{rts}}$ from the newly upsampled CASs\\
Update $\hat{\cY} \leftarrow \tilde{\cY}$, ${\cT} \leftarrow {\cT}{}^{\text{rts}}$\\
}
\KwOutput{The updated pseudo label set $\hat{\cY}$, the updated index set $\cT$ and the learned Network-OTS and Network-RTS}

\end{algorithm}

After the low-resolution CASs for each branch are generated, we then combine the generated low-resolution CASs from all branches to produce the final CAS as the ouput of the Network-RTS.
Specifically, for the $m^{\text{th}}$ branch, we temporally upsample the low-resolution CASs $\mA^{\text{rl}}_m$ to the original resolution to produce the original-resolution CASs $\tilde{\mA}_m\in R^{(C+1)\times T}$ by copying the action class scores from their neighboring frames.
The final CAS for Network-RTS $\tilde{\mA}$ is generated by averaging across all upsampled original-resolution CASs from each branch.
Similar to the Network-OTS, in the training stage, we use $\tilde{\mA}$ to generate the pseudo labels $\tilde{\vy}_t$ and the index set $\cT^{\text{rts}}$ and update $\hat{\vy}_t$ and $\cT$ by using $\tilde{\vy}_t$ and $\cT^{\text{rts}}$, respectively.
In this way, the pseudo labels generated from the Network-OTS are refined.

Let us take an $M$-branch Network-RTS ($M=1$) as an example. 
For action class $c$, given an 8-frame video with the ground-truth label $[1,1,1,1,1,1,1,1]$, and we assume the Network-OTS generates a CAS as $[0.7,0.6,0.6,0.5,0.7,0.5,0.8,0.7]$.
In contrast, if the Network-RTS reduces the temporal resolution of the input video with a scale factor of 2 and we can generate a low-resolution CAS as $[0.7,0.6,0.7,0.8]$ for the same input video, the low-resolution CAS can be temporally upsampled to the original resolution by copying from the neighboring frames and we obtain the original-resolution CAS as $[0.7,0.7,0.6,0.6,0.7,0.7,0.8,0.8]$.
When using a threshold $\theta_p=0.7$, we can generate the pseudo labels $[1,I,I,I,1,I,1,1]$ and $[1,1,I,I,1,1,1,1]$ from the Network-OTS and the Network-RTS, respectively, where $I$ represents that the corresponding samples are discarded and not used in the training process.
Inspired by the co-training strategy, which uses the predicted pseudo lables from two different views to help train better models, in our work, we exploit the complementary information between the Network-OTS and the Network-RTS by exchanging the generated pseudo labels.
For example, the refined pseudo labels output by the Network-OTS can be used to improve the Network-RTS by using them as new supervision to retrain the Network-RTS, and vice versa.
Considering the completion of the above process as one stage, our PTLR framework can be applied in a multi-stage fashion to progressively refine the pseudo labels in order to iteratively train better Network-OTS and Network-RTS.
The training process of the overall framework is summarized in Algorithm~\ref{algo:PTLR}.

\begin{table*}
\centering
\caption{Comparison (mAPs \%) on the THUMOS14 dataset when using different IoU thresholds. * denotes use of additional information (\ie, the number of action instances in videos).}
\vspace{+1mm}
\begin{tabular}{c | c | c c c c c c c c c} 
 \hline
 \multirow{2}{*}{Supervision} & \multirow{2}{*}{Methods} & \multicolumn{9}{c}{IoU threshold $\delta$}\\ 
  & & 0.1 & 0.2 & 0.3 & 0.4 & 0.5 & 0.6 & 0.7 & 0.8 & 0.9 \\ 
 \hline
 \multirow{12}{*}{Full} & SCNN~\cite{shou2016temporal} & 47.7 & 43.5 & 36.3 & 28.7 & 19.0  & 10.3 & 5.3 & - & - \\
 & REINFORCE\cite{yeung2016end} & 48.9 & 44.0 & 36.0 & 26.4 & 17.1 & - & - & - & - \\
 & CDC~\cite{shou2017cdc} & - & - & 40.1 & 29.4 & 23.3 & 13.1 & 7.9 & - & - \\
 & SMS\cite{yuan2017temporal} & 51.0 & 45.2 & 36.5 & 27.8 & 17.8 & - & - & - & - \\
 & TRUN~\cite{gao2017turn} & 54.0 & 50.9 & 44.1 & 34.9 & 25.6 & - & - & - & - \\
 & R-C3D~\cite{xu2017r} & 54.5 & 51.5 & 44.8 & 35.6 & 28.9 & - & - & - & - \\
 & SSN~\cite{zhao2017temporal} & 66.0 & 59.4 & 51.9 & 41.0 & 29.8 & - & - & - & - \\
 & BSN~\cite{lin2018bsn} & - & - & 53.5 & 45.0 & 36.9 & 28.4 & 20.0 & - & - \\ 
 & MGG~\cite{liu2019multi} & - & -& 53.9 & 46.8 & 37.4 & 29.5 & \textbf{21.3} & - & - \\
 & BMN~\cite{lin2019bmn} & - & - & 56.0 & 47.4 & 38.8 & \textbf{29.7} & 20.5 & - & - \\
 & TAL-Net~\cite{chao2018rethinking} & 59.8 & 57.1 & 53.2 & 48.5 & 42.8 & 33.8 & 20.8 & - & -\\
 & P-GCN~\cite{zeng2019graph} & \textbf{69.5} & \textbf{67.8} & \textbf{63.6} & \textbf{57.8} & \textbf{49.1} & - & - & - & -  \\
 \hline
 $\text{Weak}^{*}$ & STAR\cite{xu2019segregated} & 68.8 & 60.0 & 48.7 & 34.7 & 23.0 & - & - & - & - \\
 \hline
 \multirow{7}{*}{Weak} & STPN~\cite{nguyen2018weakly} & 52.0 & 44.7 & 35.5 & 25.8 & 16.9 & 9.9 & 4.3 & 1.2 & 0.1 \\
 & W-TALC\cite{paul2018w} & 55.2 & 49.6 & 40.1 & 31.1 & 22.8 & - & 7.6 & - & - \\
 & MAAN\cite{yuan2019marginalized} & 59.8 & 50.8 & 41.1 & 30.6 & 20.3 & 12.0 & 6.9 & 2.6 & 0.2 \\
 & CMCS\cite{liu2019completeness} & 57.4 & 50.8 & 41.2 & 32.1 & 23.1 & 15.0 & 7.0 & - & - \\
 & BM\cite{nguyen2019weakly} & 60.4 & \textbf{56.0} & 46.6 & 37.5 & 26.8 & 17.6 & 9.0 & 3.3 & 0.4 \\
 & Bas-Net\cite{lee2020background} & 58.2 & 52.0 & 44.6 & 36.0 & 27.0 & 18.6 & 10.4 & 3.9 & \textbf{0.5} \\
 & DGAM~\cite{Shi_2020_CVPR} & 60.0 & 54.2 & 46.8 & 38.2 & 28.8 & 19.8 & 11.4 & 3.6 & 0.4\\
 & Ours & \textbf{61.2} & 55.5 & \textbf{47.1} & \textbf{38.5} & \textbf{29.7} & \textbf{20.1} & \textbf{11.5} & \textbf{4.3} & 0.3\\
 \hline
\end{tabular}
\label{tab:thumos}
\end{table*}

\textbf{Inference}~~
In the inference stage, we do not apply the multi-stage strategy.
For efficient inference, only the Network-OTS and Network-RTS in the last stage are used to generate the proposals.
Specifically, for each class, we use the threshold $\theta_{cls}$ to discard the frames with their action class scores lower than $\theta_{cls}$, and the action proposals are produced by grouping the consecutive frames.
We generate the proposals from both Network-OTS and Network-RTS, and apply the contrast between the inner and the outer areas of each proposal as in~\cite{liu2019completeness} to produce the confidence score of the proposal.
To produce the final results, we combine the proposals for both Network-OTS and Network-RTS, and apply Non-Maximum Suppression (NMS) to remove the redundant proposals.


\begin{table}
\centering
\caption{Comparison (mAPs \%) on the ActivityNet v1.3 (val) dataset when using different IoU thresholds. * denotes use of additional information (\ie, the number of action instances in videos).}
\vspace{+1mm}
\begin{tabular}{c | c | c c c c } 
 \hline
 \multirow{2}{*}{Supervision} & \multirow{2}{*}{Methods} & \multicolumn{4}{c}{IoU threshold $\delta$}\\ 
 & & 0.5 & 0.75 & 0.95 & AVG \\ 
 \hline
 \multirow{6}{*}{Full} & CDC~\cite{shou2017cdc} &\textbf{ 45.30} & 26.00 & 0.20 & 23.80\\
 & TCN~\cite{dai2017temporal} & 36.44 & 21.15 & 3.90 & -\\
 & R-C3D~\cite{xu2017r} & 26.80 & - & - & -\\
 & SSN~\cite{zhao2017temporal} & 39.12 & 23.48 & \textbf{5.49 }& 23.98\\
 & TAL-Net~\cite{chao2018rethinking} & 38.23 & 18.30 & 1.30 & 20.22\\
 & P-GCN\cite{zeng2019graph} & 42.90 & \textbf{28.14} & 2.47 & \textbf{26.99}\\
 \hline
 $\text{Weak}^{*}$ & STAR\cite{xu2019segregated} & 31.1 & 18.8 & 4.7 & - \\
 \hline
 \multirow{6}{*}{Weak} & STPN\cite{nguyen2018weakly} & 29.3 & 16.9 & 2.6 & - \\
 & MAAN\cite{yuan2019marginalized} & 33.7 & 21.9 & 5.5 & - \\
 & CMCS\cite{liu2019completeness} & 34.0 & 20.9 & \textbf{5.7} & 21.2 \\
 & BM\cite{nguyen2019weakly} & 36.4 & 19.2 & 2.9 & - \\
 & Bas-Net\cite{lee2020background} & 34.5 & 22.5 & 4.9 & 22.2 \\
 & Ours & \textbf{36.9} & \textbf{24.1} & 5.6 & \textbf{23.7}\\
 \hline
\end{tabular}
\label{tab:activitynet}
\end{table}

\begin{table*}
\centering
\caption{MAPs (\%) of our Initial Label Generation method and the variants of our method without using the cross-stream consistency loss and the temporal multi-resolution consistency loss on the THUMOS14 dataset.}
\vspace{+1mm}
\begin{tabular}{c | c c | c c c c c c c c c } 
 \hline
 & Cross-stream & Temporal multi-resolution & \multicolumn{9}{c}{IoU threshold $\delta$}\\ 
 & Consistency Loss & Consistency Loss & 0.1 & 0.2 & 0.3 & 0.4 &0.5 & 0.6 & 0.7 & 0.8 & 0.9 \\ 
 \hline
 ILG w/o cs \& tm & \xmark & \xmark & 57.5 & 51.2 & 43.7 & 35.5 & 26.2 & 17.9 & 10.1 & 3.8 & \textbf{0.5}\\
 ILG w/o tm& \cmark & \xmark & 58.3 & 52.1 & 44.5 & 36.1 & 26.5 & 18.7 & 10.5 & 3.5 & 0.3 \\
 ILG w/o cs& \xmark & \cmark & 58.7 & 52.9 & \textbf{45.7} & 36.5 & 27.1 & 18.5 & 10.7 & 3.9 & 0.3 \\
 ILG& \cmark & \cmark & \textbf{59.1} & \textbf{53.2} & 45.5 & \textbf{37.1} & \textbf{27.9} & \textbf{19.2} & \textbf{11.6} & \textbf{4.2} & 0.4\\
 \hline
\end{tabular}
\label{tab:loss}
\end{table*}

\begin{table*}[]
    \centering
    \caption{Results (precision (\%) and recall rate (\%)) of our Initial Label Generation method and the variants of our method without using the cross-stream consistency loss and the temporal multi-resolution consistency loss on the THUMOS14 dataset.}
    \begin{tabular}{c| c c | c c}
    \hline
    & Cross-stream & Temporal multi-resolution & \multirow{2}{*}{Precision (\%)} & \multirow{2}{*}{Recall Rate (\%)}\\ 
    & Consistency Loss & Consistency Loss &  & \\ 
    \hline
    ILG w/o cs \& tm &\xmark & \xmark & 81.2 & 23.4 \\
    ILG w/o tm& \cmark & \xmark & 79.8 & 29.1 \\
    ILG w/o cs& \xmark & \cmark & \textbf{81.9} & 30.5 \\
    ILG& \cmark & \cmark & 80.5 & \textbf{37.4}\\
    \hline
    \end{tabular}
    \label{tab:plabel}
\end{table*}

\section{Experiments}

In this section, we first introduce our experimental setup and datasets in Section~\ref{sec:setup}, and then compare our method with state-of-the-art methods in Section~\ref{sec:comparision}.
Finally, we conduct ablation study in Section~\ref{sec:ablation}.

\subsection{Experiment Setup}
\label{sec:setup}

\textbf{Datasets}~~
We evaluation our weakly supervised temporal action localization method on two datasets: THUMOS14~\cite{jiang2014thumos} and ActivitiNet v1.3~\cite{caba2015activitynet}. It is noteworthy that we only use the video-level labels during the training process and the temporal annotation is used merely for evaluation.

\emph{-THUMOS14} consists of 1,010 and 1,574 untrimmed videos with 20 action classes in the validation and test sets, respectively.
However, only 200 videos in the validation set and 212 videos in the test set are temporally annotated and each of them contains at least one action instance.
We use these 200 videos in the validation set as our training samples, and the 212 videos in the testing set as our testing samples.

\emph{-ActivityNet v1.3} contains 19,994 videos from 200 different activities.
The whole video set is divided into training, validation and testing sets with the ratio of 2:1:1.
We train our model on the training set and evaluate our model based on the validation set as the annotation for the testing set is not publicly available.

\textbf{Implementation details}~~
We use the I3D network~\cite{carreira2017quo}, which is pre-trained based on the Kineics~\cite{carreira2017quo} dataset, as our feature extractor to extract both the RGB and optical flow features. 
The optical flow maps are generated by using the TVL1 algorithm~\cite{wedel2009improved}. 
For each video, instead of extracting the features for every frame, we divide the video into a set of non-overlapping 16-frame segments and extract the RGB and flow features for each segment. 
In this way, we can reduce the computational cost.

We implement our framework by using PyTorch~\cite{paszke2017automatic}. We train our Initial Label Generation (ILG) module with the learning rate of 0.0001 on the THUMOS14 dataset and 0.0005 on the ActivityNet v1.3 dataset for 1500 iterations.
For our Progressive Temporal Label Refinement (PTLR) framework, we use 3 branches in the Network-RTS. 
We train both Network-OTS and Network-RTS with the learning rates of 0.001 and 0.002 on the THUMOS14 and the ActivityNet v1.3 datasets, respectively.
We also set $\theta_p$ as 0.7, 0.65, 0.6, and 0.55 for stage 1, stage 2, stage 3, and stage 4 to generate pseudo labels in our PTLR framework, respectively.
Note that our method achieve the best results when using three stages.
For inference, we follow~\cite{lee2020background} to use a set of thresholds from 0 to 0.5 with the interval of 0.025 and perform NMS with the threshold of 0.7 to remove highly overlapped proposals.
All the experiments are conducted on a single GTX 1080Ti GPU.

\textbf{Evaluation metrics}~~
Following the existing works~\cite{lee2020background,nguyen2019weakly}, we use mean average precision (mAP) to evaluate our method for weakly supervised temporal action localization.
An action proposal is considered as correct if its overlap with the ground-truth instance is larger than a threshold $\delta$ and the action class labels are correctly predicted.
We also compute the intersection-over-union (IoU) score to measure the overlap between action proposals and ground-truth instances.

To evaluate the quality of the generated pseudo labels, we also calculate the precision and recall for the generated pseudo labels.
The precision is calculated by computing the ratio of the number of selected frames with correct pseudo labels over the total number of selected frames with pseudo labels.
To calculate the recall, we compute the ratio of the number of frames with correct pseudo labels over the total number of frames in all video.


\subsection{Comparision with the State-of-the-art Methods}
\label{sec:comparision}

We compare our proposed method with the state-of-the-art fully-supervised and weakly-supervised approaches at different IoU thresholds.
All results of the existing methods are quoted from their original works.
Among the weakly supervised methods, it is noteworthy that STAR~\cite{xu2019segregated} is specially denoted as ``$\text{weak}^{*}$" since it uses the number of action instances in each video as additional information during the training process.

\begin{table*}
    \centering
    \caption{Comparison (mAP(\%) of three PTLR-related methods at different number of stages when using different IoU thresholds on the THUMOS dataset}
    \begin{tabular}{c | c | c c c c c c c c c}
         \hline
         \multirow{2}{*}{Methods} & \multirow{2}{*}{Stage} & \multicolumn{9}{c}{IoU threshold $\delta$} \\
         & & 0.1 & 0.2 & 0.3 & 0.4 & 0.5 & 0.6 & 0.7 & 0.8 & 0.9 \\
         \hline
         \multirow{4}{*}{PTLR-OTS} & 1 & 57.7 & 51.9 & 44.5 & 36.9 & 27.5 & 18.3 & 10.2 & 4.2 & \textbf{0.4} \\
         & 2 & 58.2 & 52.6 & \textbf{45.1} & \textbf{37.2} & 27.8 & \textbf{18.6} & \textbf{10.5} & \textbf{4.3} & \textbf{0.4} \\
         & 3 & \textbf{58.9} & \textbf{52.9} & 44.8 & 36.9 & \textbf{27.9} & 18.2 & 10.4 & 4.2 & 0.3 \\
         & 4 & 58.5 & 52.6 & 44.9 & 36.5 & 27.4 & 18.3 & 10.3 & 4.2 & 0.3 \\ 
         \hline
         \multirow{4}{*}{PTLR-RTS} & 1 & 58.7 & 52.8 & 45.1 & 37.3 & 27.9 & 18.9 & 10.3 & 4.2 & 0.4 \\
         & 2 & 59.5 & \textbf{53.7} & 45.9 & 37.6 & \textbf{28.3} & \textbf{19.3} & \textbf{10.5} & 4.1 & \textbf{0.5} \\
         & 3 & \textbf{59.8} & 53.6 & \textbf{46.2} & \textbf{37.7} & 27.9 & 19.1 & 10.2 & 4.1 & 0.4 \\
         & 4 & 59.3 & 53.1 & 45.7 & 37.3 & 27.7 & 19.0 & 10.2 & \textbf{4.3} & 0.4 \\
         \hline
         \multirow{4}{*}{PTLR} & 1 & 59.5 & 53.6 & 45.9 & 37.7 & 28.5 & 19.5 & 10.7 & 4.1 & 0.3 \\
         & 2 & 60.7 & 54.9 & 46.5 & 38.2 & 29.1 & 20.0 & 11.2 & \textbf{4.5} & 0.3 \\
         & 3 & \textbf{61.2} & \textbf{55.5} & \textbf{47.1} & \textbf{38.5} & \textbf{29.7} & \textbf{20.1} & \textbf{11.5} & 4.3 & 0.3 \\
         & 4 & 60.1 & 55.1 & 46.9 & 38.2 & 29.2 & 19.8 & 11.3 & 4.1 & \textbf{0.4} \\  
         \hline
    \end{tabular}
    \label{tab:ptlr}
\end{table*}

\begin{table}[]
    \centering
    \caption{Comparison of the pseudo labels generated by the Network-OTS and the Network-RTS in our PTLR framework at different number of stages in terms of precision (\%) and recall rate (\%) on the THUMOS14 dataset. }
    \begin{tabular}{c | c | c | c }
        \hline
         Stage & Network & Precision (\%) & Recall  \\
         \hline
         \multirow{2}{*}{1} & Network-OST & 81.4 & 42.2 \\
         & Network-RTS & 82.6 & 49.6 \\
         \hline
         \multirow{2}{*}{2} & Network-OST & 78.9 & 55.1 \\
         & Network-RTS & 79.2 & 61.3 \\
         \hline
         \multirow{2}{*}{3} & Network-OST & 74.5 & 63.1 \\
         & Network-RTS & 73.2 & 67.4 \\
         \hline
         \multirow{2}{*}{4} & Network-OST & 70.1 & 68.9 \\
         & Network-RTS & 69.7 & 70.3\\
         \hline
    \end{tabular}
    \label{tab:plable-ptlr}
\end{table}

\textbf{THUMOS14}~~
We report the mAP results on the THUMOS14 dataset in Table~\ref{tab:thumos}. Our method outperforms the existing methods that use only video-level labels as weak supervision at all IoU thresholds except $\delta=0.2$ and $\delta=0.9$.
Note that our method is built upon Bas-Net~\cite{lee2020background} with the additional cross-stream consistency loss and temporal multi-resolution consistency loss as the extra constraints to generate class activation sequences (CASs) and a new PTLR module to refine the pseudo labels.
When $\delta=0.5$, our method can achieve the mAP of 29.7\%, which is 2.7\% higher than that of the baseline Bas-Net.
We believe this improvement is due to multi-resolution information exploitation in our ILG and PTLR modules.
It is also observed that STAR~\cite{xu2019segregated} outperforms our method when $\delta \leq 0.3$.
As mentioned, STAR~\cite{xu2019segregated} used additional weak information when training the model. 
Even so, our method still can achieve better mAP results when $\delta > 0.3$.
When comparing our method with the state-of-the-art fully supervised approaches, our method even outperforms several existing fully supervised approaches~\cite{xu2017r,dai2017temporal,yuan2017temporal}, while our method uses much weaker supervision.

\textbf{ActivitiNet v1.3}~~
In Table~\ref{tab:activitynet}, we report the mAPs of our method and compare different approaches at different IoU thresholds $\delta=\{0.5, 0.75, 0.95\}$ on the validation set of the ActivityNet v1.3 dataset. 
Besides, we also report the average mAP, which is calculated by averaging all mAPs across multiple IoU thresholds $\delta$ ranging from 0.5 to 0.95 with an interval of 0.05.
When compared with the existing approaches using only video-level labels during the training process, the mAPs of our method are generally the best at different IoU thresholds. 
It is only slightly worse than CMCS~\cite{liu2019completeness} when $\delta$=0.95.
Moreover, in terms of the average mAP, our method achieves the mAP of 23.7\%, which is 2.5\% and 1.5\% higher than that of CMCS~\cite{liu2019completeness} and our baseline method Bas-Net, respectively, which demonstrates the effectiveness of our proposed method.
It it noteworthy that, STAR~\cite{xu2019segregated} takes advantage of extra weak information, however, it is still worse than our method at all IoU thresholds on the ActivityNet v1.3 dataset.


\begin{figure*}[!t]
\includegraphics[width=\linewidth]{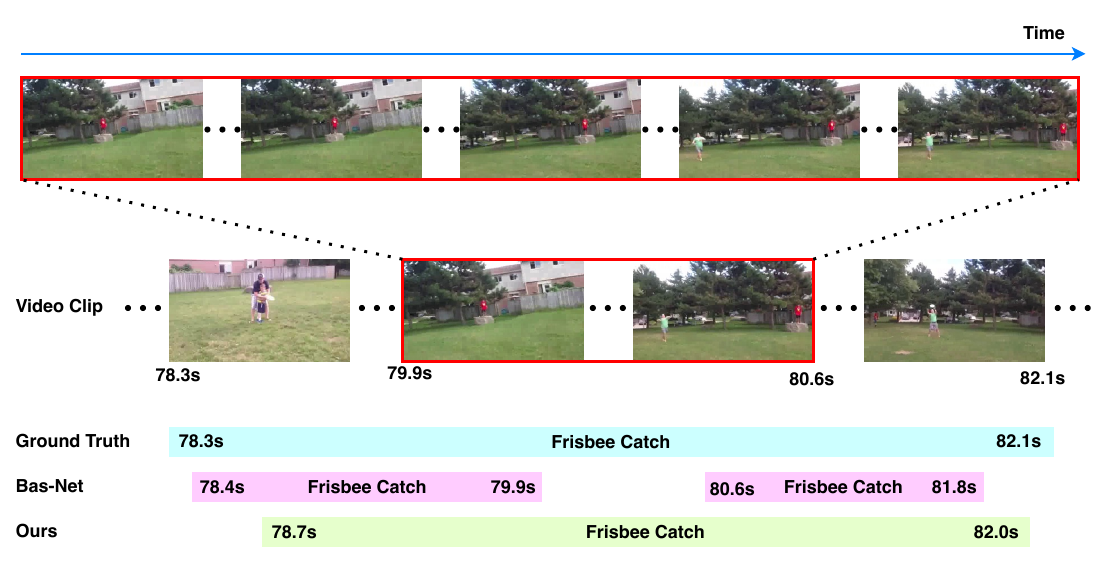}
\caption{Qualitative comparison of the  results from the baseline method Bas-Net~\cite{lee2020background} and our method on the THUMOS14 dataset. The true action instance of the action class ``Frisbee Catch" starts at the 78.3-th second and ends at the 81.8-th second of the video clip. In this example, our method can successfully localize the ground-truth action instance, while the Bas-Net method only detects the discriminative periods (\ie, from the 78.4-th second to the 79.9-th second, and from the 80.6-th second to the 81.8-th second) and misses the less discriminative period (\ie, from the 79.9-th second to the 80.6-th second) of the action instance. More frames during the period between the 79.9-th second and the 80.6-th second of the video clip are shown in the red boxes.} 
\label{fig:visualization}
\end{figure*}

\subsection{Ablation Study}
\label{sec:ablation}

In this section, we take THUMOS14 as an example to conduct ablation study and investigate the contributions of different components in our proposed method.

\textbf{Cross-stream consistency loss and temporal multi-resolution consistency loss}~~
In order to investigate the contributions of our cross-stream consistency loss and temporal multi-resolution consistency loss in our ILG method, we compare our complete ILG method with three simplified versions: (1) ILG w/o cs \& tm: we remove both the cross-stream consistency loss and the temporal multi-resolution consistency loss from our complete ILG module; (2) ILG w/o tm: we remove the temporal multi-resolution consistency loss from our complete ILG module; (3) ILG w/o cs: we remove the cross-stream consistency loss from our ILG module.
It is noting that our baseline method ILG w/o cs \& tm combines two stream information in a late fusion way, which is different from the baseline method Bas-Net~\cite{lee2020background}.
We compare the quality of the CASs generated by these three ILG-related variants and our complete ILG method.
In Table~\ref{tab:loss}, we follow~\cite{lee2020background} to produce the action proposals from the CASs generated by these four ILG-related methods and report the mAPs at different IoU thresholds.
From Table~\ref{tab:loss}, we observe that our complete ILG method can achieve the best temporal action localization performance among all four ILG-related methods, which demonstrates that both our cross-stream consistency loss and temporal multi-resolution consistency loss can help improve the quality of the generated CASs.

We also compare the quality of the initial pseudo labels generated by these four ILG-related methods.
Table~\ref{tab:plabel} reports the precision and the recall rates of the initial pseudo labels from each of them.
When compared with the other three counterparts, although the precision of the initial pseudo labels generated by our complete ILG method is slightly lower when compared with the other three methods, it can generate the initial pseudo labels with much higher recall rate.
The results indicate that more correctly labelled training samples are provided to train the Network-OTS in our PTLR framework.
These results also demonstrate the effectiveness of our cross-stream consistency loss and temporal multi-resolution consistency loss.

\textbf{Progressive temporal label refinement framework}~~
We compare our complete PTLR framework with two PTLR-related variants: (1) in PTLR-OTS: we remove the Network-RTS from our complete PTLR framework, and use the Network-OTS to iteratively refine the pseudo labels and retrain the network itself; (2) in PTLR-RTS: we remove the Network-OTS from our complete PTLR framework, and use the Network-RTS to iteratively refine the pseudo labels and retrain the network itself.
We use the initial pseudo labels generated from our complete ILG module as supervision to train these three PTLR-related methods, and report the mAP at different stages when using different IoU thresholds in Table~\ref{tab:ptlr}.
We observe that the mAPs of our PTLR method at all IoU thresholds are significantly improved as the number of stages increases, which demonstrates that our proposed method can progressively enhance the temporal action localization performance.
It is noted that our PTLR achieves the best mAP results after three stages.
In terms of mAPs, our PTLR outperforms PTLR-OTS and PTLR-RTS at every stage across almost all IoU thresholds, which denmonstrates the effectiveness of our cooperative pseudo label refinement strategy for improving the performance of the Network-OTS and the Network-RTS.

In Table~\ref{tab:plable-ptlr}, we report the precision and the recall rate of the pseudo labels generated by Network-OTS and Network-RTS in our PTLR framework.
For both Network-OTS and Network-RTS, we observe that the precision of the pseudo labels decreases as the number of stages increases.
A possible explanation is we reduce the threshold $\theta_p$ to select confident samples and add more training samples at each stage, which increases the possibility of assigning wrong action class labels to the selected training samples.
At each stage, the recall rate of the Network-RTS is significantly higher than that of the Network-OTS, which indicates that our Network-RTS can successfully refine the pseudo labels and add back the previously discarded frames with reliable pseudo labels.


\subsection{Qualitative Results}

In addition to the quantitative performance comparison, the qualitative results are also provided to demonstrate the performance of our method.
In Fig.~\ref{fig:visualization}, we show the visualization results obtained from the baseline method Bas-Net~\cite{lee2020background} and our method.
We observe that the Bas-Net method fails to detect the less disriminative period of the ground-truth action instance, and thus divides a complete action instance into two falsely detected action segments.
Meanwhile, our method can successfully detect a complete action segment that covers most period (including the less discriminative period) of the ground-truth action instance, which demonstrates the effectiveness of our proposed method.

\section{Conclusion}

In this paper, we have proposed an Initial Label Generation (ILG) module to take advantage of the cross-stream and the temporal multi-resolution consistency constraints to generate the frame-level pseudo action class labels by only using video-level annotation when training our network.
We have also proposed a Progressive Temporal Label Refinement (PTLR) framework consisting of two networks called Network-OTS and Network-RTS, which iterateively refines the pseudo labels to benefit the network training process.
Comprehensive experiments on both THUMOS14 and ActivityNet v1.3 datasets demonstrate the effectiveness of our proposed method.


%





\ifCLASSOPTIONcaptionsoff
  \newpage
\fi



%
{\small
\bibliographystyle{IEEEtran}
\bibliography{egbib}
}

%








\end{document}